\documentclass[letterpaper, 10 pt, conference]{ieeeconf}  % Comment this line out if you need a4paper

\IEEEoverridecommandlockouts                              % This command is only needed if 
\usepackage{amsmath} % assumes amsmath package installed
\usepackage{amssymb}  % assumes amsmath package installed
\usepackage{color}

\usepackage{graphicx}
\usepackage{bbold}
\usepackage{multirow}
\usepackage{subcaption}
\usepackage{url}
\usepackage{tikz}

\newcommand\submittedtext{%
  \footnotesize This work has been submitted to the IEEE for possible publication. Copyright may be transferred without notice, after which this version may no longer be accessible.}
  
\newcommand\submittednotice{%
\begin{tikzpicture}[remember picture,overlay]
\node[anchor=south,yshift=10pt] at (current page.south) {\fbox{\parbox{\dimexpr\textwidth-\fboxsep-\fboxrule\relax}{\submittedtext}}};
\end{tikzpicture}%
}

\newcommand{\bE}{\begin{eqnarray*}}
  \newcommand{\eE}{\end{eqnarray*}}
  \newcommand{\bEn}{\begin{eqnarray}}
  \newcommand{\eEn}{\end{eqnarray}}

\newcommand{\x}{\mathbf{x}}

\newcommand{\Det}[1]{\text{det}\!\left(#1\right)}

\title{\LARGE \bf
Global Optimization of Stochastic Black-Box Functions with Arbitrary Noise Distributions using Wilson Score Kernel Density Estimation
}

\author{Thorbjørn Mosekjær Iversen, Lars Carøe Sørensen, Simon Faarvang Mathiesen, Henrik Gordon Petersen
\thanks{All authers are with SDU Robotics, The Maersk Mc-Kinney Moller Institute, University of Southern {\tt\small \{thmi,lcs,simat,hgp\}@mmmi.sdu.dk}}%
}% <-this % stops a space

\begin{document}

\maketitle
\submittednotice
\thispagestyle{empty}
\pagestyle{empty}

%%%%%%%%%%%%%%%%%%%%%%%%%%%%%%%%%%%%%%%%%%%%%%%%%%%%%%%%%%%%%%%%%%%%%%%%%%%%%%%%

\begin{abstract}

Many optimization problems in robotics involve the optimization of time-expensive black-box functions, such as those involving complex simulations or evaluation of real-world experiments. Furthermore, these functions are often stochastic as repeated experiments are subject to unmeasurable disturbances. Bayesian optimization can be used to optimize such methods in an efficient manner by deploying a probabilistic function estimator to estimate with a given confidence so that regions of the search space can be pruned away. Consequently, the success of the Bayesian optimization depends on the function estimator's ability to provide informative confidence bounds. Existing function estimators require many function evaluations to infer the underlying confidence or depend on modeling of the disturbances. In this paper, it is shown that the confidence bounds provided by the Wilson Score Kernel Density Estimator (WS-KDE) are applicable as excellent bounds to any stochastic function with an output confined to the closed interval [0;1] regardless of the distribution of the output. This finding opens up the use of WS-KDE for stable global optimization on a wider range of cost functions. The properties of WS-KDE in the context of Bayesian optimization are demonstrated in simulation and applied to the problem of automated trap design for vibrational part feeders.
\end{abstract}
\section{INTRODUCTION}

Simulations have always been at the core of robotics research. As computational resources increase, so does the complexity of the simulations. Complex dynamic robot simulations often have no analytical form, are expensive to evaluate, and have no easy way to estimate the derivatives. One approach to perform optimization using such simulations is the use of Bayesian optimization. This approach to optimization seeks to use estimates of the confidence bounds on the optimization parameters to infer which areas of parameter space are likely to contain the optimal solution, and which can be excluded from the search space.

The estimation of confidence bounds in Bayesian optimization is often done using Gaussian Processes (GPs), which rely on the already evaluated data points and chosen covariance function, which models the correlation between different points in the parameter space. If the function being optimized is stochastic, the distribution of the output must also be estimated, which is a challenge as the distribution of the output variable is typically unknown. If sufficient experiments are made for a specific parameter set, the normal approximation is applicable. However, the point of utilizing Bayesian optimization is to reduce the number of needed computations, making the usability of the Gaussian Process reliant on the user's estimate of the output distribution.

When the stochastic function is binary, Wilson Score Kernel Density Estimation (WS-KDE) provides a theoretically sound alternative to Gaussian Processes. WS-KDE combines kernel density estimation with the Wilson Score approximation for the success probability of a binary function, which is accurate even with a low number of trials. This has allowed for Bayesian optimization of trap parameters in the context of automating the design of mechanical vibratory feeder traps, where the outcome of each expensive dynamic simulation is categorized as a success or failure~\cite{sorensen2020wilson}. 

\begin{figure}
    \centering
\includegraphics[width=\linewidth]{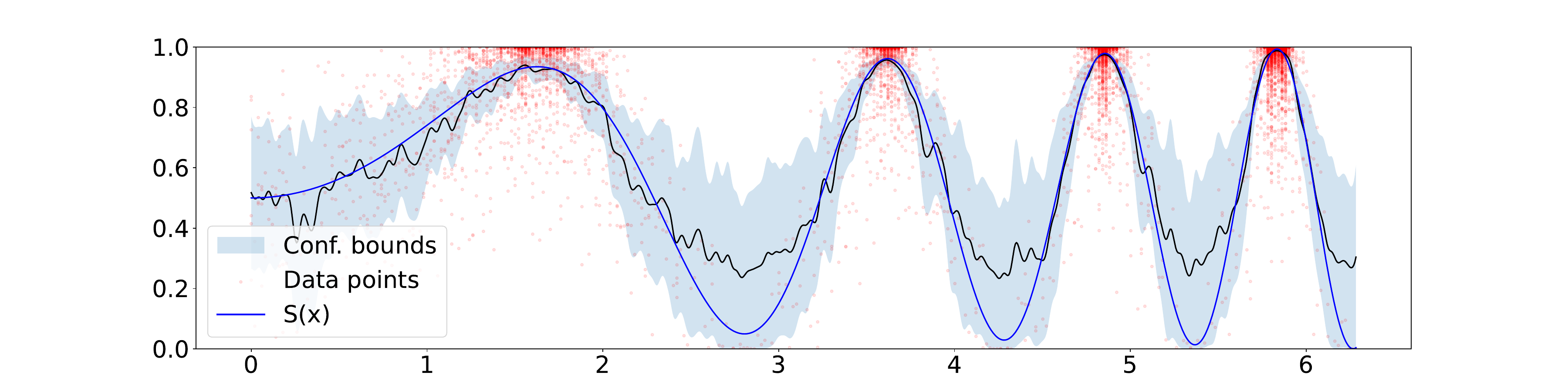}
    \caption{The figure shows the Wilson Score Kernel Density Estimator's function estimate after 10000 Bayesian optimization iterations. The confidence bounds indicate that the troughs have been pruned away early in the process letting the optimization focus on sampling around the potential maxima. While the noise in the example is Beta distributed, the confidence bounds are excellent for any distribution with support on the closed interval [0;1].}
    \label{fig:frontFigure}
\end{figure}

In this work, we prove that the bounds provided by WS-KDE are conservative bounds on the expectation value of an arbitrary stochastic function bounded on the closed interval [0;1], regardless of the output's distribution and without having to rely on the normal approximation. This enables confidence bounds to be applicable even for regions with very few or even no samples. Through simulations, the coverage of the confidence bounds is investigated, as well as the convergence of a Bayesian optimization process using WS-KDE as function estimator.

To demonstrate the applicability of our method to a real optimization problem, we use WS-KDE-based Bayesian optimization to automate the design of a vibratory part feeder trap. A vibratory part feeder~\cite{Boothroyd2005} is a class of part-feeding devices widely used in assembly automation. These feeders convey a bulk of parts past orientation devices known as traps which either by the principle of reorientation or rejection and recirculation, ensure that all parts are delivered to the subsequent automation system in the same desired position and orientation. The method of conveying is by controlled vibration which leads to a system with highly stochastic behavior even though parts under conveying generally will jump between a finite set of stable poses. Designing these feeders hence requires a significant amount of data points to obtain precise estimates of the trap sequence' impact on throughput and accuracy of the resulting part pose. Previous work~\cite{mathiesen2018optimisation} has demonstrated that simulation-based optimization can be effective for trap design. However, a designer must have a great level of domain knowledge to estimate which stable poses of a part pair well with a specific trap principle in order to avoid resorting to an exhaustive list of optimization tasks when searching for the optimal trap. We show that the use of a new non-binary metric reduces the complexity for the feeder designer as it removes the requirement of defining a desired binary outcome for the optimization task.

Python and C++ implementations of the multivariate WS-KDE equations are provided at \url{https://github.com/SDU-Robotics/wilson-score-kde} making it easy to apply our method to data-driven optimization problems for robotic system components.
% Generally applicable C++ and Python implementations of the multivariate WS-KDE equations will be provided, making it easy to apply our method to data-driven optimization problems for robotic system components.

The paper is structured as follows: Section~\ref{sec:relatedWork} provides an overview of related work. Section~\ref{sec:method} provides a brief overview of the Wilson Score confidence bounds, the WS-KDE, and our proof that these provide conservative bounds to arbitrary stochastic functions with an output confined to [0;1]. Section~\ref{sec:experiments} describes the simulation and the experiments and presents the results. Finally, Section~\ref{sec:Conclusion} concludes the paper and outlines suggestions for future work.
\section{Related Work}
\label{sec:relatedWork}
Bayesian optimization is a well-established method for optimizing expensive black-box functions, where sample efficiency in the optimization is essential. The method relies on an iterative cycle of using existing data to infer a stochastic model of the cost function, which in turn is used to determine the next point to sample~\cite{garnett2023bayesian}.

The literature on function estimation is extensive with multiple different approaches to estimate probabilistic functions, including Bayesian neural networks (BNNs), GPs, and kernel density estimators.

BNNs estimate the posterior distribution of the weights, allowing for sampling-based estimation of the mean and variance of the network outputs~\cite{jospin2022hands}. However, as neural networks rely on high quantities of training data, they are ill-suited in the context of Bayesian optimization.

GPs are often used as function estimators in the context of Bayesian optimization. The simplest form assumes no noise on the measurement or the noise is assumed to be Gaussian~\cite{pml1Book}. In cases where the noise is not Gaussian, common alternatives are to use heavy-tailed distributions, such as Student's t-distribution, or use mixture models if the distribution is multi-modal~\cite{rasmussen2003gaussian}. Notably, the predictions of a GP are highly dependent on the chosen prior and Kernel hyperparameters. Furthermore, GPs have a time complexity of $O(N^3)$ which limits their use with large data samples unless approximations are employed~\cite{he2023bi}.

While kernel density estimation does not explicitly estimate a function, it estimates densities which are used to infer the mean and variance at a given point based on the sampled points in its proximity~\cite{hardle2004nonparametric}. Consequently, kernel density estimators do not impose a specific functional form on the function estimates. The time complexity of estimating the mean and variance for a single point is $O(N)$ making it able to handle larger datasets, although as a kernel-based method, it is sensitive to the curse of dimensions.

Wilson Score Kernel Density Estimation, which is the method our work is based on, was presented in~\cite{sorensen2020wilson}. In it, the authors observed that standard kernel density estimation, where the normal approximation is used to infer the mean and variance~\cite{hardle2004nonparametric}, provides poor estimates of the confidence bounds for the success rate for Bernoulli trials for a low number of samples. The work of~\cite{sorensen2020wilson} combines kernel density estimation with Wilson Score confidence intervals~\cite{agresti1998approximate}, which provide accurate confidence bounds even for few samples.

To the best of our knowledge, no other function estimator can provide confidence bounds on arbitrary output distributions confined to [0;1], with only few observations, without assuming a prior distribution of the noise or relying on data-intensive sampling.

% 17.2.3 Comparison to kernel regression \cite{pml1Book}

\section{Method}
\label{sec:method}

This section first rigidly defines the optimization problem, followed by a brief review of the Wilson Score confidence interval, and our proof of the Wilson Score confidence interval's applicability to arbitrary stochastic functions with support on the interval [0;1]. Finally, we provide a summary of Wilson Score Kernel density estimation.

\subsection{Problem definition}
\label{subsec:problemdef}
The focus of this paper is the optimization of control parameters $x$ for a computationally expensive black-box function $f(x,P)\to[0;1]$ where $x\in B$ is the control parameter to be optimized and $B\subset {\mathbb{R}}^n$ is a closed and bounded set chosen by the user. The parameter $P$ is a random perturbation belonging to an unknown distribution $\rho(P)$. This parameter models the impact of random perturbations in the function evaluations of the individual experiments. Hence $f(x,P)$ is a deterministic function, but $f$ is stochastic when viewed as a function of the control parameter $x$ only.

In this paper, we present a method for a data-driven derivation of the control parameter
$\hat{x}\in B$ that optimizes the expectation value $S(x)$ of $f$. This expectation value can be formally expressed as: 
\begin{align}
S(x) \equiv \int f(x, P)\rho(P)dP    
\end{align}
We allow f(x,P) to be discontinuous in both x and P, but our method will exploit the reasonable assumption that S(x) is continuous so that Kernel Density Estimation can be applied. 
% The optimal control parameters can now be expressed as the parameters that finds the control parameter $x^\in B$ that optimizes the expectation value $S(x)$ of $f$. This expectation value can be formally expressed as: 
% \begin{align}
% S(x) \equiv \int f(x, P)\rho(P)dP    
% \end{align}
% We allow f(x,P) to be discontinuous in both x and P, but we make the reasonable assumption that S(x) is continuous. 
%The optimal control parameters can now be expressed as the parameters that optimize the expectation value of $f$ under the given set of control parameters:

% \begin{align}
% \hat{x} = \arg\max_x S(x)
% \end{align}
% where
% \begin{align}
% S(x) &= \mathbb{E}_{P\sim\rho(P)}[f(x,P)]\\
%  &= \int f(x,P)\rho(P)dP
% \end{align}

Although the distribution of perturbations is unknown, using a data-driven approach to sample from $\rho(P)$ allows $S(x)$ to be estimated by vast repetitions of real-world experiments with the same $x$ and unknown perturbations $P_i$:

\begin{align}S(x)=\lim_{n\rightarrow \infty}\frac{1}{n}\sum_{i=1}^n f(x,P_i)
\end{align} 

Repeated sampling of $f$ in this way is of course impractical as we will only estimate $S(x)$ for a single value of $x$. Hence, this approach must be combined with a method that updates the value of $x$ based on previous results and by this search for $\hat{x}$. For the updating, it is important that the size of the regions of parameter space, which can contain $\hat{x}$ are reduced quickly to be able to securely and efficiently prune the search space. Bayesian optimization, which has been shown to work well in this context, follows an iterative cycle of estimating $S(x)$ with associated confidence $\sigma(x)$ based on previously sampled points and then using some selection criteria to determine the next point to be evaluated. To prune areas of $x$ which are unlikely to contain $\hat{x}$ without risk of pruning the interesting regions, it is essential to have an accurate conservative estimation of the confidence bounds.

A function estimator should for any $x_i\in B$ at any stage in the experimental sequence provide an estimate of the mean $p(x_i)$ with an associated confidence interval 
\begin{equation}
CI(x_i)= \Bigg[ p(x_i) \pm z \sqrt{\sigma^2(x_i)} \Bigg] \, ,
 \label{eq:true:confideninterval}
\end{equation}
where $z$ is defined as the $(1-\frac{\alpha}{2})$ quantile for a two-sided interval. Regions of $x_i$'s where the upper bound of $CI(x_i)$ is smaller than the current highest lower bound can then be pruned.

\subsection{Wilson Score confidence interval}
The Normal Approximation, which provides the basis of the standard kernel density estimator, is based on the assumption that for a given $x_i$ we have performed $n_i$ experiments. This allows us to approximate the confidence interval using the sample mean and variance:
\bEn
CI_{na}(x_i)=\hat{p}_{na}(x_i)\pm \frac{z}{n_{i}}\sqrt{\sum^{n_{i}}_{j=1}\left(y_{j}-\hat{p}_{na}(x_i)\right)^2}
\eEn
where $\hat{p}_{na}(x_i)=\frac{1}{n_{i}} \sum^{n_{i}}_{j=1} y_{j}$, and $y_j$ is the outcome of the $j$'th experiment

For $x_i$'s with no samples, this expression makes no sense and for low sample sizes, the confidence intervals of the Normal Approximation is far too optimistic. To account for this, Wilson suggested to rephrase the computation of the confidence interval. The starting point is to have an estimate $\hat p$ of the unknown true mean $p$, and a desired confidence $z_\alpha$. Hence, the value of the true mean $p$ must reside between the solutions to
\bEn
\label{eq:pW}
(\hat{p}-p)^2=z^2_{\alpha/2}\sigma^2
\eEn
where $\sigma^2$ is the unknown true variance at $x_i$. This equation unfortunately has two unknown parameters $p$ and $\sigma^2$, but Wilson considered this for an underlying binomial distribution where the mean was obtained from $n$ experiments, and hence $\sigma^2=\frac1{n}p(1-p)$. Insertion of this expression into (\ref{eq:pW}) and solving for $p$ yields
\bEn
\label{eq:WCI}
\hat{p}_{ws}-z_{\alpha/2}\hat\sigma_{ws}\leq p\leq \hat{p}_{ws}+z_{\alpha/2}\hat\sigma_{ws}
\eEn
where
\begin{align}
 \hat{p}_{ws} = \frac{n}{n+z_{\alpha/2}^2} \hat{p} + \frac{1}{2 n} z_{\alpha/2}^2 \, ,
 \label{eq:bernoulli:ws:mean}
\end{align}
and
\begin{align}
\hat \sigma_{ws} = \frac{n}{n+z_{\alpha/2}^2}z_{\alpha/2}\sqrt{\frac{1}{n}\hat{p}(1-\hat{p}) +\frac{z_{\alpha/2}^2}{4n^2}} \, ,
 \label{eq:bernoulli:ws:variance}
\end{align}
The expressions in (\ref{eq:WCI})--(\ref{eq:bernoulli:ws:variance}) yields a statistically robust and conservative estimate for the CI even with few or no experiments. 

\subsection{Wilson score applied to arbitrary distributions with support on [0;1]}
We now consider $n$ experiments taken from a general probability density with support on the interval $[0;1]$. If the true mean is $p$, it is then a remarkably simple observation that the true variance satisfies that
\bEn
\sigma^2\leq \frac1{n}p(1-p)
\eEn 
as the variance can never be larger than the variance from a binomial distribution with the same $p$.
Hence, we may still apply (\ref{eq:bernoulli:ws:mean}) and (\ref{eq:bernoulli:ws:variance}) for the mean and variance, albeit they are more conservative than for the binomial distribution. Notice that the term $\frac{1}{n}\hat{p}(1-\hat{p})$ should not be replaced by the observed variance $\hat\sigma^2$ unless experiments are Bernoulli trials. The Wilson score estimate is in this generalized case hence only depending on the observed mean $\hat{p}$ and the sample size $n$.

\subsection{Wilson score applied with Kernel Density Estimation}
To achieve fast convergence towards $\hat{x}$ for the search method, it is also important to exploit that $S(x)$ will be a continuous function over most of $B$. Kernel Density Estimation (KDE) \cite{hardle2004nonparametric} is an easy-to-use method to exploit this. In KDE the sample mean is estimated as

% The first step in Kernel Density Estimation (KDE) is to define a smoothing kernel $K_{h,X}(x)$ centered around $X$. we then define a density of experiments, $f(x)$, at an arbitrary parameter set, $x$. This estimate is defined as:
% \begin{align}
%  \hat{f}_h(x) = \frac{1}{n} \sum_{i=1}^{n} K_{h,x_i}(x) \, ,
%  \label{eq:kde:density:density}
% \end{align}
% where there was an experiment at the $x_i$'s. If the value of the $i$'th experiment is $y_i$, we get the Kernel Density estimate of the mean
\bEn
\hat{m}_h(x) = \frac{\hat{f}_{h,Y}(x)}{\hat{f}_h(x)} = \frac{ \sum_{i=1}^{n} K_{h,x_i}(x) \, y_i}{\sum_{i=1}^{n} K_{h,x_i}(x)} 
\eEn
where $K_{h,X}(x)$ is a smoothing kernel centered around $X$ and $n$ is the total set of experiments done until now.
The KDE-weighted number of experiments at $x$ is computed as
% In Kernel Density Estimation, a formula for a variance estimate also exists, but as we wish to link KDE to Wilson scores, this estimate can be omitted as it is not used for the Wilson score. Instead, we explicitly state the number of experiments at $x$ after KDE smoothing found in LC S\o rensen et al:
\bEn
n_h(x)=\frac{nh}{\|K\|_2^2}\hat{f}_h(x)
\eEn
where
\bEn
\|K\|_2^2=\int K(u)^2du
\eEn
is the squared $L_2$ norm of the identity kernel. In \cite{hardle2004nonparametric}, a formula for a variance estimate based on the Normal Approximation was also derived. However, here we deploy the result from \cite{sorensen2020wilson} where the Wilson Score estimate was used in the context of kernel density estimation for Bernoulli trials, to the more general case of arbitrary distributions on the interval $[0;1]$. By extending the arguments from the previous section, it is easily seen that the Bernoulli trial confidence bounds also hold here. These are
\begin{align}
 \hat{p}_{ws,kde} = \frac{n_h(x)}{n_h(x)+z_{\alpha/2}^2} \hat{m}_h(x) + \frac{1}{2 n_h(x)} z_{\alpha/2}^2 \, ,
 \label{eq:kde:mean}
\end{align}
and
\bEn
&&\hat \sigma_{ws,kde}= \frac{n_h(x)}{n_h(x)+z_{\alpha/2}^2}z_{\alpha/2}\nonumber\\
&\times&\sqrt{\frac{1}{n_h(x)}\hat{m}_h(x)(1-\hat{m}_h(x)) +\frac{z_{\alpha/2}^2}{4n_h(x)^2}}  
 \label{eq:kde:variance}
\eEn

%\subsection{WS-KDE in the context of BO with arbitrary distribiutino}

%\thmi{Add argument that the conservative WS bounds are still applicable, and why conservative bounds are ok to have in a Baysian optimization setting.}
\section{Experiments}
\label{sec:experiments}

To evaluate our method we both performed simulations on a known analytic function and applied our method to a real problem. The simulation experiments provide insights into the WS-KDE function estimator's ability to estimate valid confidence bounds, and its performance when applied as function estimator as part of a Bayesian Optimization scheme. The use of WS-KDE is demonstrated on a real problem of optimizing a trap for a mechanical vibratory feeder, using a novel loss function allowing for more autonomy in the trap design, while still using the stability of the WS-KDE optimizer.

\subsection{Simulation}
\label{subsec:simulation}

Since the extension of WS-KDE to multiple dimensions is completely trivial, all simulation experiments are done on a one-dimensional function, since this allows for ease of explanation without losing generality.\footnote{The extension to multiple dimensions simply increases the computational complexity, as the required number of samples will increase exponentially with the dimensionality of the search space.}
%The primary concern moving to multiple dimensions is computational complexity and memory consumption, which is a general problem when doing global optimization of multi-dimensional black-box functions.

For the simulation experiments, we choose the following analytic function (see Fig.~\ref{fig:frontFigure}) as the ground truth expectation value $S(x)$. 
\begin{align}
    S(x) = 0.5 \left( \sin\left(a x ^ 2\right)e^{- b (2 \pi - x)} + 1\right)
    \label{eq:testfunction}
\end{align}
where $a=0.6$ and $b=0.03$.

This specific function was chosen as it contains four distinct peaks, where each subsequent peak is slightly narrower and has a slightly higher maximum than the previous. This makes it a challenging function to optimize since the slight difference in peak height is easily lost in the sampling noise.

For binomial experiments, this means that the function being sampled is $f_\text{Bern}(x) \sim \text{Bernoulli(S(x))}$. For the other experiments, we chose the stochastic function to be a beta distribution, i.e. $f_\text{beta}(x) \sim \frac{1}{\mathbf{B(\alpha,\beta)}}x^{\alpha-1}(1-x)^{\beta-1}$. We chose a beta distribution as it is a normalized probability distribution bounded on [0;1] with a closed form expectation value of $\frac{\alpha}{\alpha + \beta}$. The distribution can thus be defined by defining $S(x)$ as the expectation value and choosing a value for $\alpha\! +\! \beta$ which will control the shape of the distribution.

% The chosen $S(x)$ has four distinct peaks, where each subsequent peak is slightly narrower and has a slightly higher maximum than the previous. This makes it a challenging function to optimize since the slight difference in peak height is easily lost in the sampling noise.

\subsubsection{Properties of WS-KDE}
\begin{figure}
    \centering
    \includegraphics[width=\linewidth]{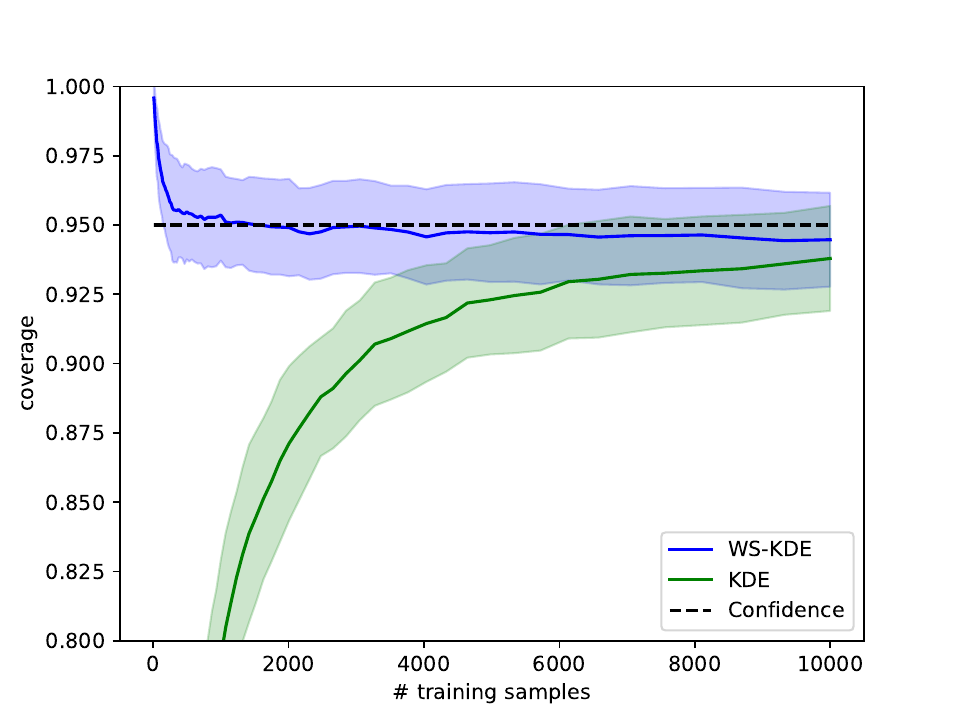}
    \caption{Comparison of the coverage of standard KDE CI's with WS-KDE CI's. As indicated, the confidence was set to 95\%.}
    \label{fig:coverage}
\end{figure}

 We first study the robustness of the confidence interval by performing experiments up to $n=10000$ to investigate the estimation bias as the number of samples increases. We tesselate the search space into a grid of n=1001 equally spaced points. For each point, we check if the ground truth expectation value is within the estimated confidence bounds
 \bEn
 \text{cov} = \frac{1}{N}\sum _{i=1}^N\mathbb{1}(x_i)
 \eEn
 where $\mathbb{1}(x_i)$ is an indicator function with a value of 1 if and only if $m(x_i)-\sigma(x_i) < S(x_i) < m(x_i)+\sigma(x_i)$. Since the simulated coverage will depend on the random sampling of the function, we repeat the experiment 100 times and plot the mean coverage with the bands denoting the mean coverage $\pm$ one standard deviation. The coverage plots for WS-KDE and KDE are shown in Fig.~ \ref{fig:coverage}. It can be seen that, as expected, the WS-KDE is significantly more accurate for low sample sizes and as the number of samples becomes large, the two coverages become more similar. Notice also that when the number of experiments becomes large, the coverage becomes less than the set confidence. This is due to the Kernel bias and will be discussed in Section \ref{sec:Conclusion}.

\begin{figure}
   \centering
   \includegraphics[width=\linewidth]{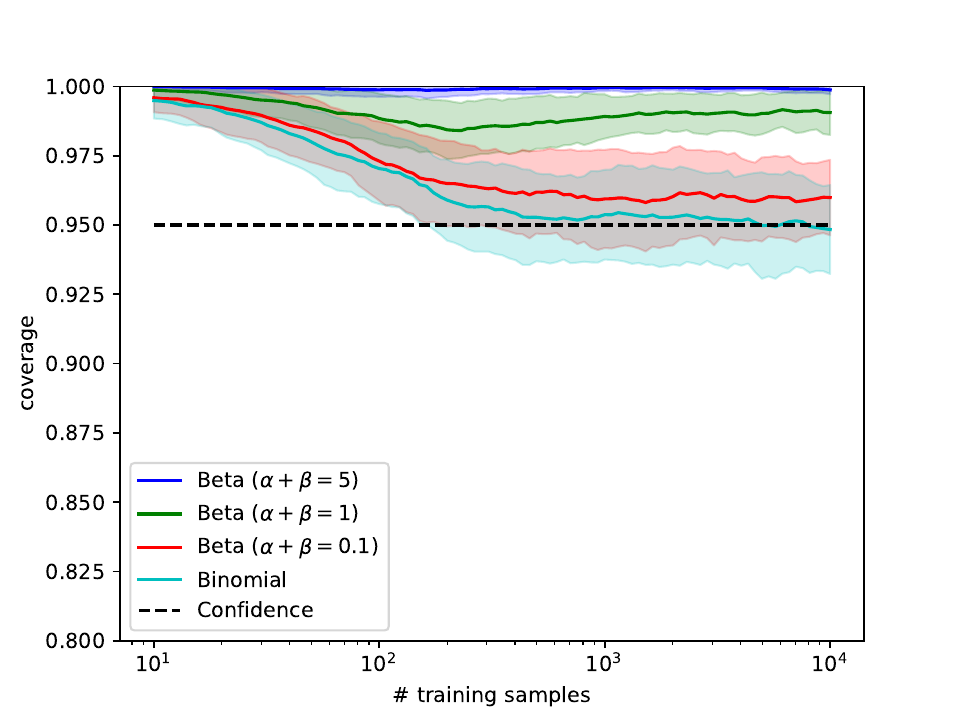}
   \caption{Comparison of the coverage of WS-KDE when applied to the analytic function (\ref{eq:testfunction}) with beta or binomially distributed noise. The shape of the beta distributions is defined by $\alpha\!+\!\beta$.}
   \label{fig:beta}
\end{figure}

The final test on the coverage of the WS-KDE was made for three beta functions. By choosing appropriate values of $\alpha\! +\! \beta$, the distribution changes from one akin to a binomial distribution to something closer to a Gaussian distribution. As expected from the proof given in the method section, all three beta functions have higher coverage than the binomial distribution (see Fig. \ref{fig:beta}).

\subsection{Evaluation of WS-KDE in the context of Bayesian optimization} \label{sec:evalWSKDE}
When used as function estimator in a Bayesian Optimization scheme, the coverage is important as it allows for informed pruning of areas of the search space that are known to not contain the maximum with a certain confidence. As the standard KDE relies on the normal approximation to estimate the confidence bounds, its estimates are unreliable for a low number of data points. Consequently pruning based on confidence can be critical as a few unlikely samples can lead to a poor function estimate with estimated high confidence. This can lead to pruning of areas containing maxima, despite the lack of statistically significant evidence. We demonstrate the stability of Bayesian optimization using WS-KDE by optimizing $S(x)$ both when the noise is binomial and when it is beta distributed with $\alpha\! +\! \beta=5$.

We present three metrics in the evaluation. The first is the maximum lower confidence bound
\begin{align}
    \text{LCB}_\text{max} = \text{max}_x \left\{m(x)-\sigma(x)\right\}  \label{eq:lcbmax}
\end{align}

This bound provides, at the statistical confidence level, the lowest function value that the optimization will lead to. Assuming adequate coverage of the function estimator, this value should generally increase with increasing number of samples. This metric is specifically interesting because it allows for early stopping when the optimization reaches a required performance. In the figures, the $\text{LCB}_\text{max}$ value is presented as a fraction of the ground truth function max $S_\text{max}$.  The second and third metrics are the total pruning rate and the false pruning rate. 
\begin{align}
    I_\text{tot} = \frac{N_\text{pruned}}{N}, \quad I_\text{false} = \frac{N_\text{false}}{N}
\end{align}
where $N_\text{pruned}$ is the number of x-values for which $m(x)+\sigma(x) < LCB_\text{max}$, and $N_\text{false}$ is the number of pruned x-values where the ground truth expectation value is above the threshold, i.e. $S(x)> LCB_\text{max}$. Preferably the total pruning rate should be as high as possible to limit the search space, while the false pruning rate should be close to zero.

\begin{figure}
    \centering
    \begin{subfigure}[t]{1\linewidth}
        \centering
        \includegraphics[width=0.49\linewidth]{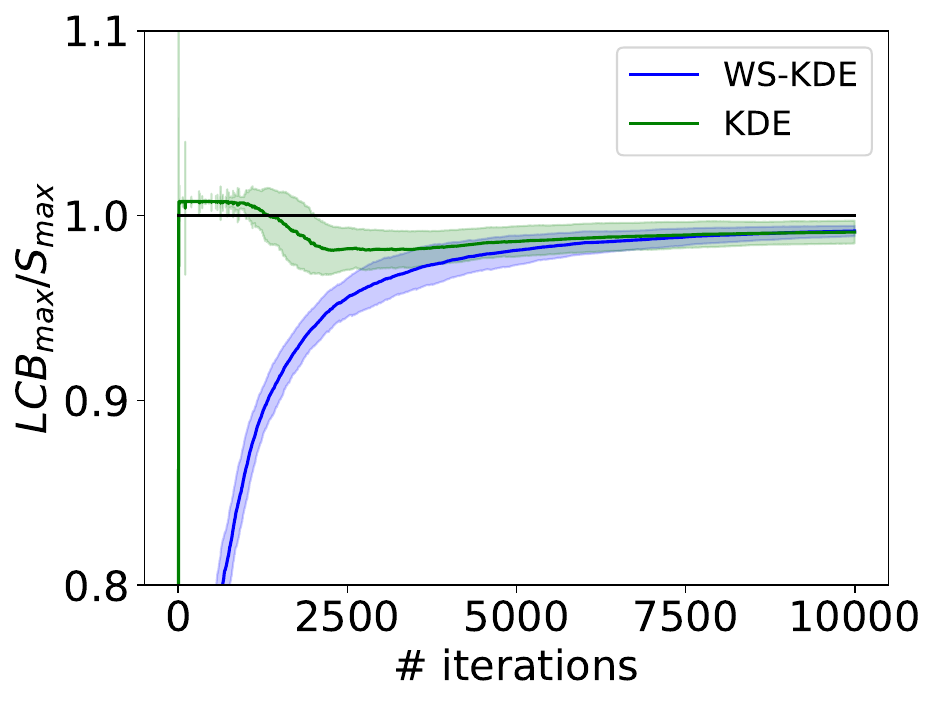}
        \includegraphics[width=0.49\linewidth]{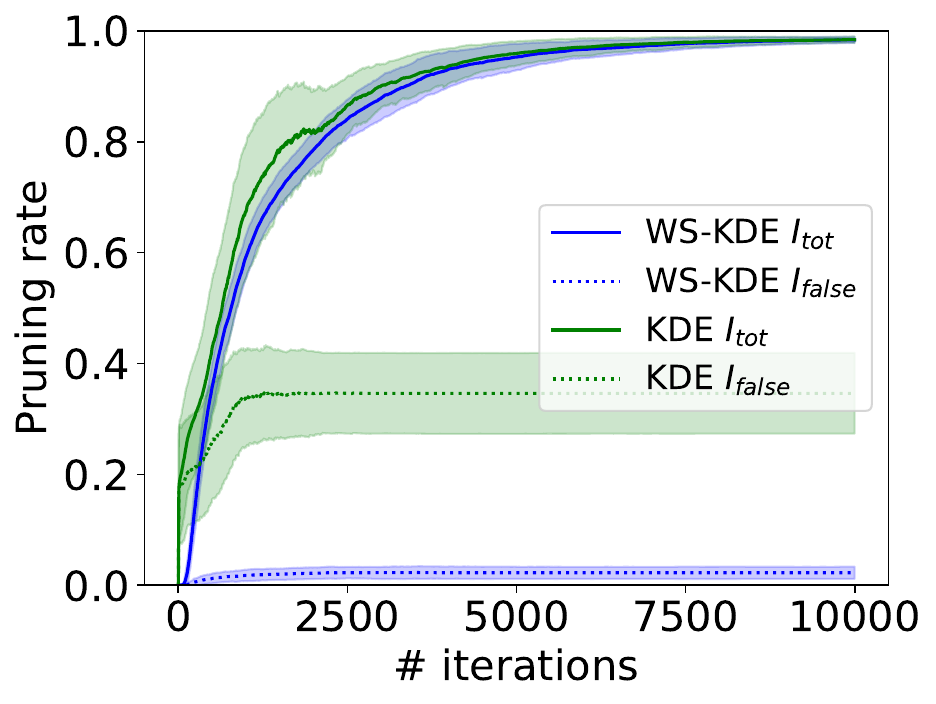}
        \caption{Binomial}
    \end{subfigure}
    \begin{subfigure}[t]{\linewidth}
        \centering
        \includegraphics[width=0.49\linewidth]{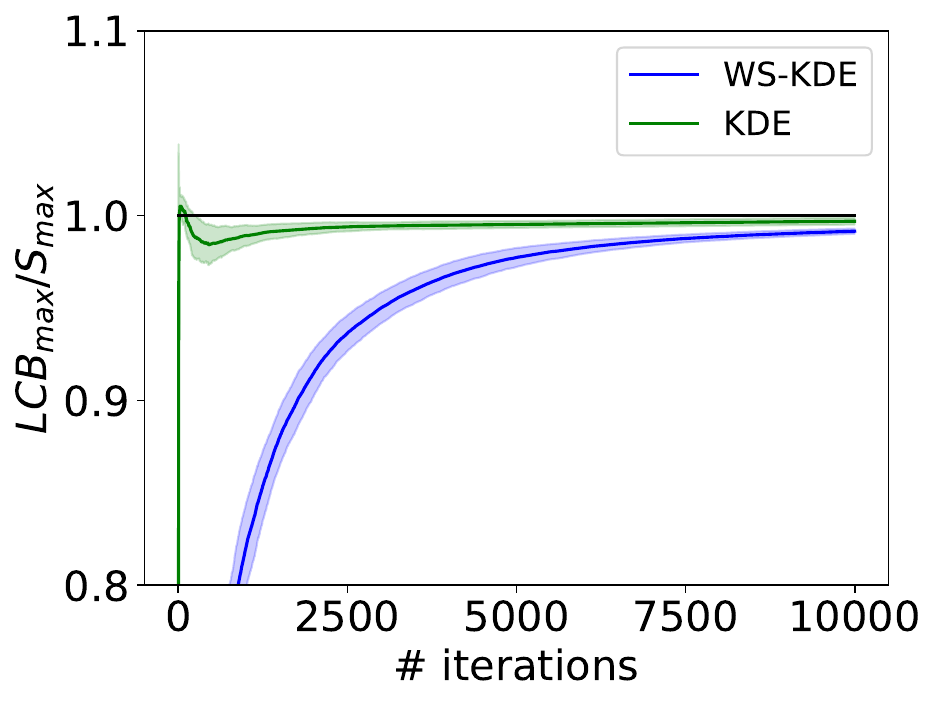}
        \includegraphics[width=0.49\linewidth]{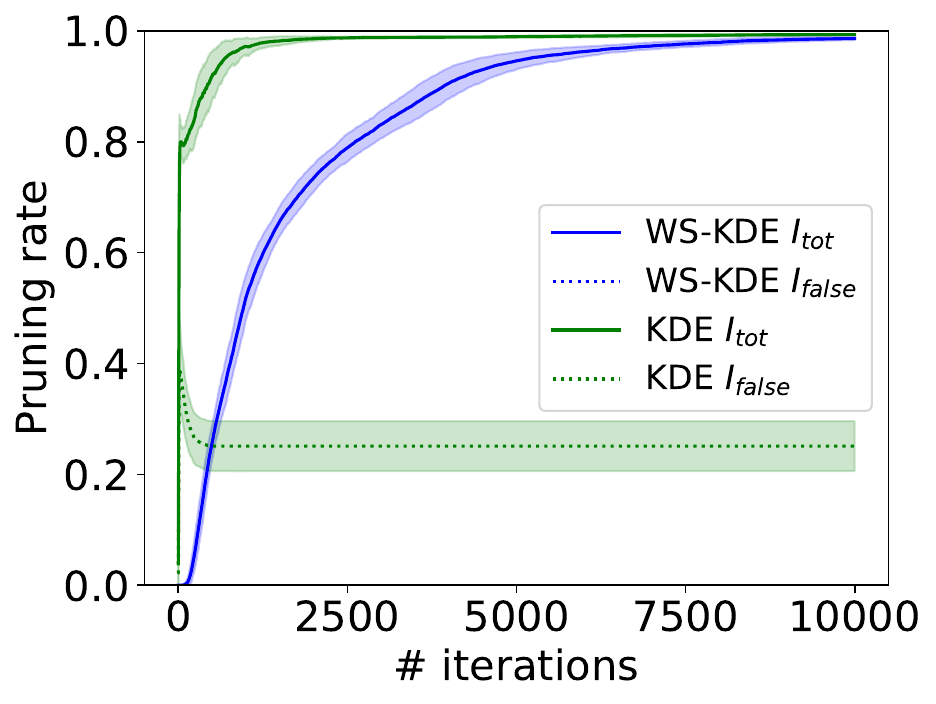}
        \caption{Beta}
    \end{subfigure}
    \caption{Comparison of standard KDE and WS-KDE for 10000 Bayesian optimization iterations of the test function (\ref{eq:testfunction}) with respectively binomial and beta distributed noise ($\alpha\!+\!\beta=5$). Left: The highest lower confidence bound as a fraction of the ground truth function maximum, Right: Fraction of pruned $x_i$'s and fraction of falsely pruned $x_i$'s.}
    \label{fig:pruning}
\end{figure}

We use pruning as global exclusion and select the next sample point using uniform sampling from the non-pruned search space. To perform the pruning and sampling of a new search point, we use a grid search with a grid spacing approximately equal to the kernel width $h$. With the choice of kernel $h=0.02$ and parameter space $[0,2\pi]$ we thus sample 315 points. We use the pruning scheme mentioned in Sec. \ref{subsec:problemdef} where we prune $x_i$’s where the upper confidence bound of $x_i$ is less than the current highest lower bound.

The results of the evaluation with binomial noise and beta noise are shown in Fig. \ref{fig:pruning}a and \ref{fig:pruning}b. The results show that for both the binomial and beta experiment, standard KDE is over-optimistic within the first $\sim 1000$ iterations, after which it converges to a more realistic estimate as enough samples are available. This instability in the estimate of the $\text{LCB}_\text{max}$ can as mentioned lead to early pruning of the search space that may contain the global maxima. This can be seen on the right in Fig. \ref{fig:pruning} which shows that the total pruning rate is high in the beginning, at the cost of a high false pruning rate. On the contrary, WS-KDE converges almost monotonically towards the correct maximum. While the increase in total pruning rate is slower for WS-KDE, the false pruning rate is very low for the binomial experiment, and almost non-existent for the beta experiment due to 
the high coverage of the beta function.

\begin{table}
\renewcommand{\arraystretch}{1.3}
    \caption{Results of 100 experiments searching for the global maximum. The columns correspond to the four peaks from left to right indicating the three local maxima ($L1,L2,L3$) and the global maximum ($GM$).}
    \label{tab:peak}
    \centering
    \begin{tabular}{ll|cccc}
    & & \multicolumn{4}{c}{Local or Global maximum}\\
    \multicolumn{2}{c|}{Experiment} & L1 & L2 & L3 & GM \\
    \hline
    \multirow{2}{*}{Binomial} &  KDE   & 0 & 1&10 &89\\
      & WS-KDE & 0& 0& 0 &\textbf{100} \\
      \hline
       \multirow{2}{*}{Beta ($\alpha\! +\! \beta=5$)} &   KDE  &  0&0 &1 &99\\
       & WS-KDE & 0&0 & 0 & \textbf{100}\\
    \end{tabular}
\end{table}

To determine if the global maximum is found, we determine which peak a given optimization procedure converged to and present this in Table~\ref{tab:peak}. Again we observe instability when using standard KDE, where WS-KDE converged to the correct maximum in all experiments.

\begin{figure*}
    \centering
    \begin{subfigure}[t]{0.49\textwidth}
        \centering
        \includegraphics[width=\linewidth]{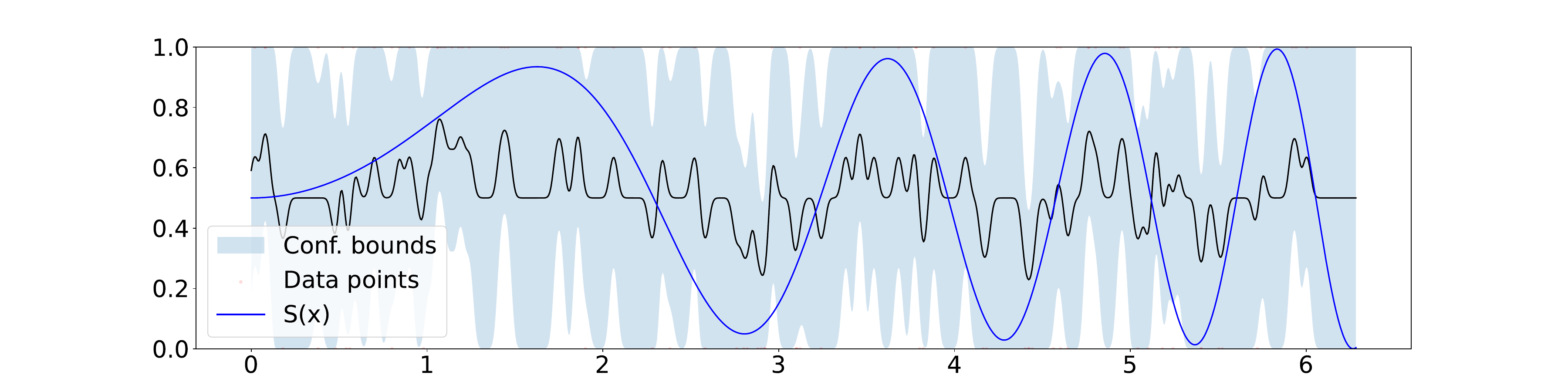}
        \includegraphics[width=\linewidth]{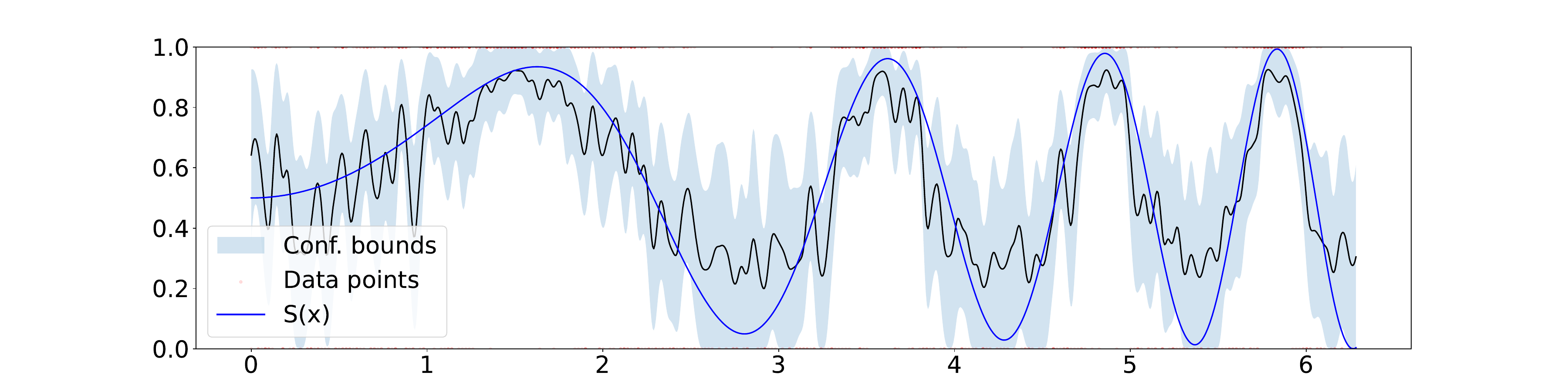}
        \includegraphics[width=\linewidth]{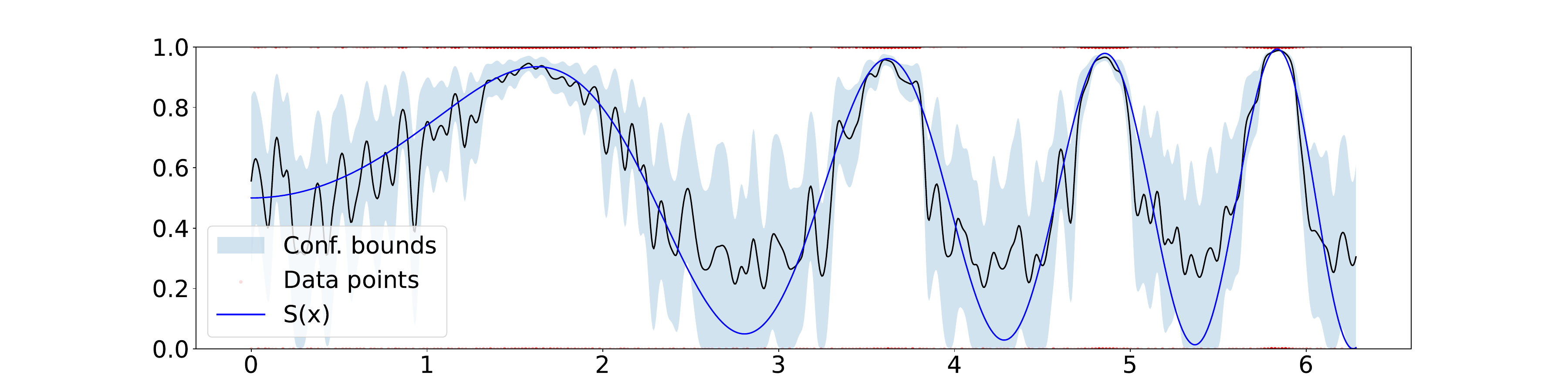}
        \caption{WS-KDE - binomial distribution}
    \end{subfigure}
    \begin{subfigure}[t]{0.49\textwidth}
        \centering
        \includegraphics[width=\linewidth]{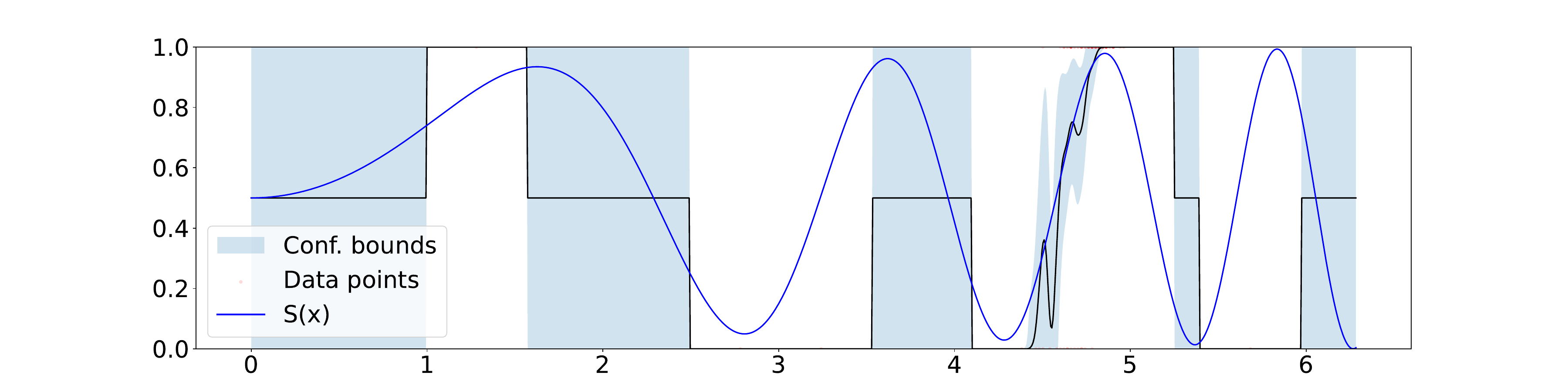}
        \includegraphics[width=\linewidth]{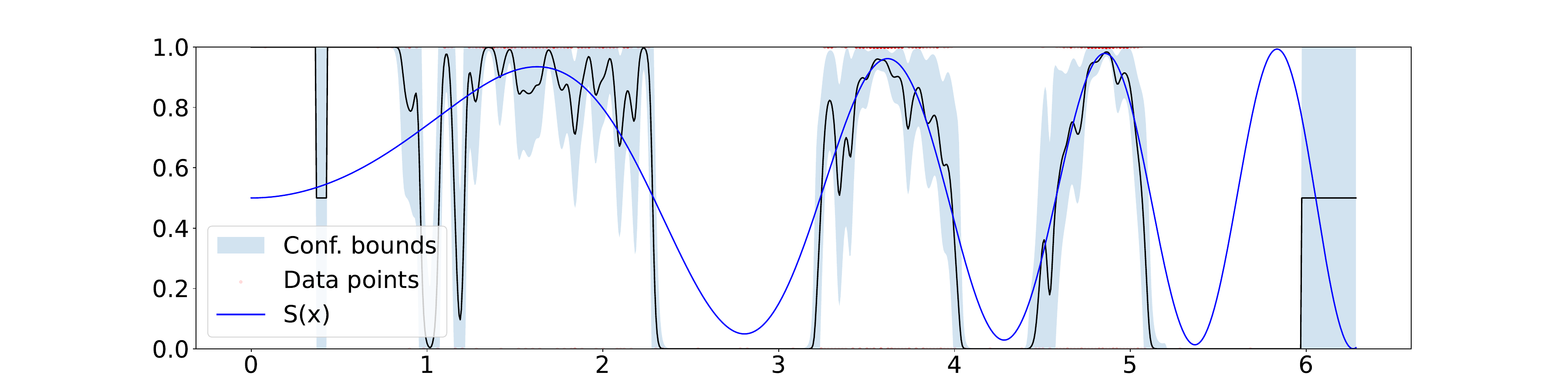}
        \includegraphics[width=\linewidth]{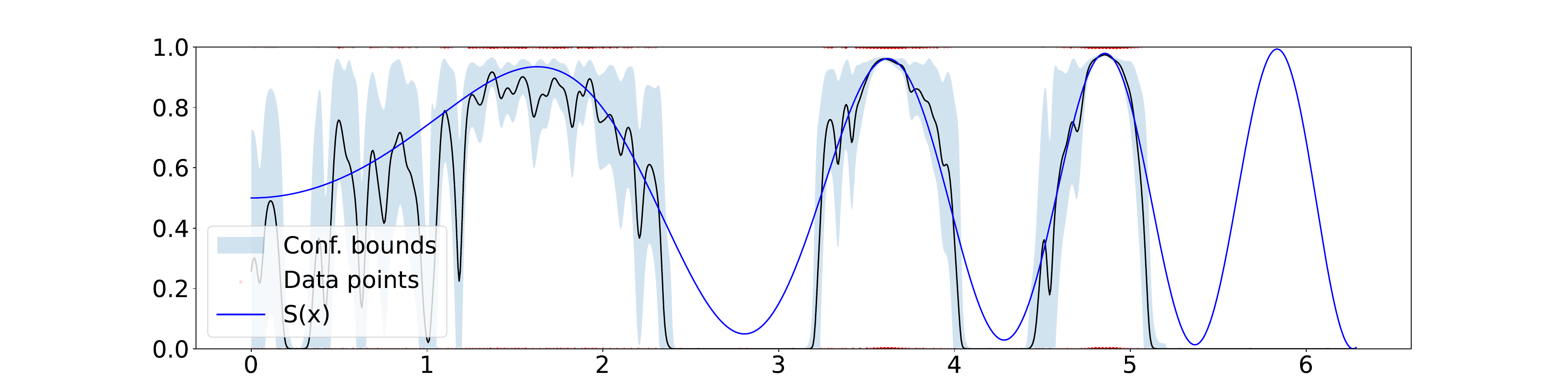}
        \caption{KDE - binomial distribution}
    \end{subfigure}

    \begin{subfigure}[t]{0.49\textwidth}
        \centering
        \includegraphics[width=\linewidth]{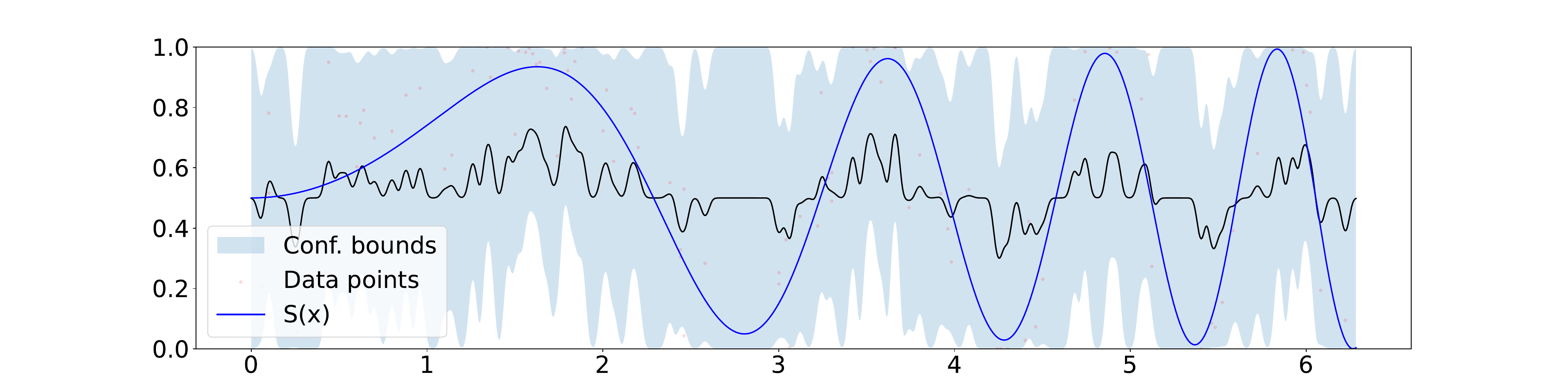}
        \includegraphics[width=\linewidth]{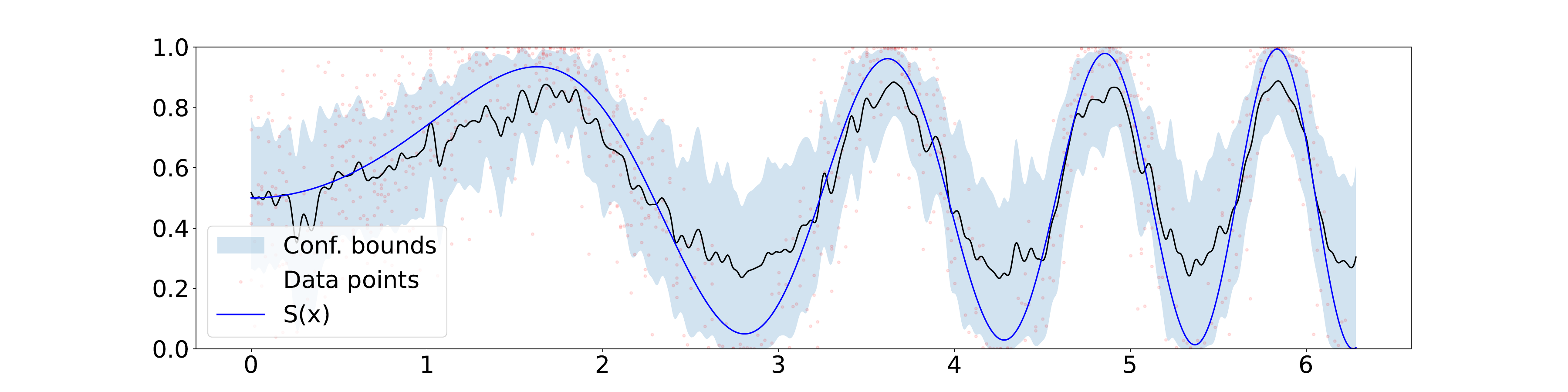}
        \includegraphics[width=\linewidth]{figures/feb26_final/func_ws_beta_h002_k0_9999_i66.pdf}
        \caption{WS-KDE - beta distribution ($\alpha\! +\! \beta=5$)}
    \end{subfigure}
    \begin{subfigure}[t]{0.49\textwidth}
        \centering
        \includegraphics[width=\linewidth]{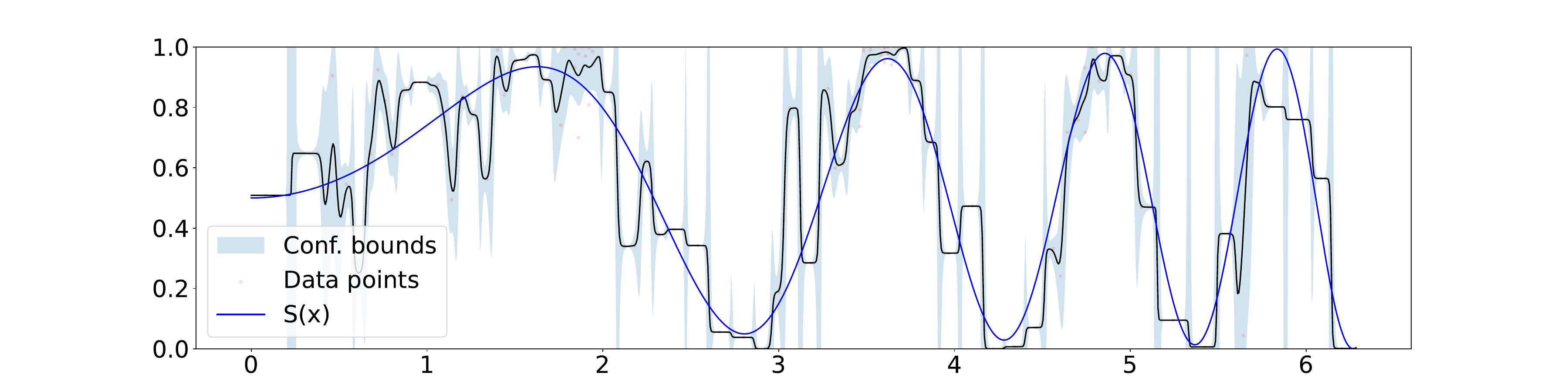}
        \includegraphics[width=\linewidth]{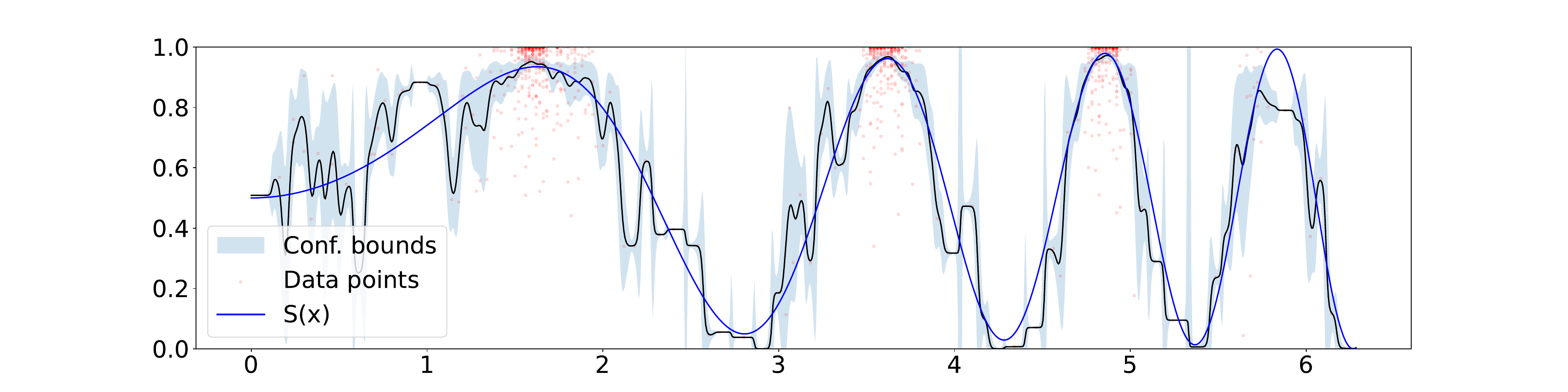}
        \includegraphics[width=\linewidth]{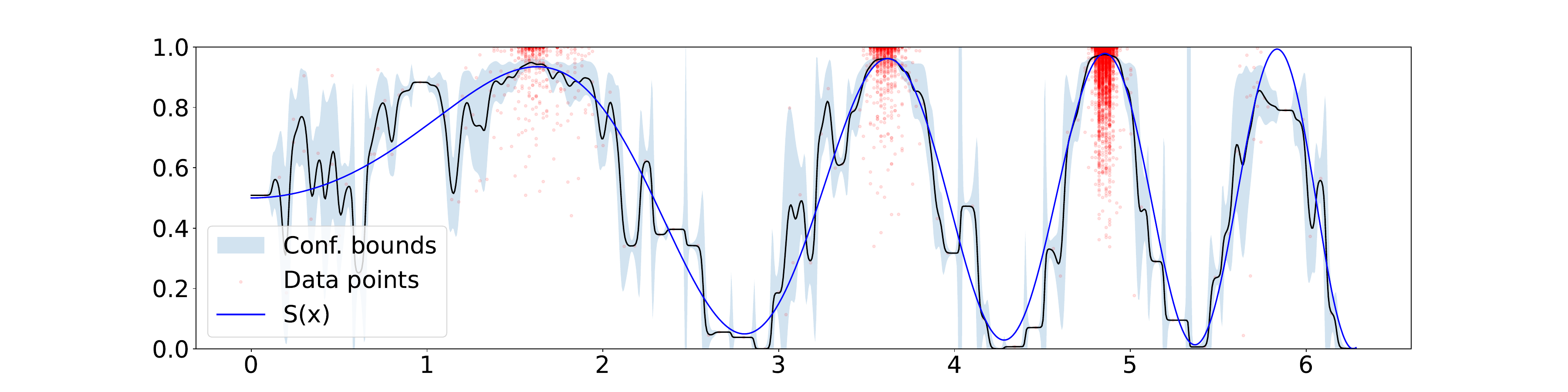}
        \caption{KDE - beta distribution ($\alpha\! +\! \beta=5$)}
    \end{subfigure}

    \caption{Visualization of the 100th, 1000th, and 10000th Bayesian optimization iteration using WS-KDE (a,c) or KDE (b,d) as function estimator for binomial (a,b) or beta (c,d) distributed function output. The optimizations are chosen where KDE fails to find the maximum, to illustrate how incorrect confidence estimates can lead to pruning of the global maximum.}
    \label{fig:qualitative}
\end{figure*}

To give insight into how WS-KDE provides the demonstrated stability, we choose optimization experiments where KDE failed to converge and visualize the function estimates after 100, 1000, and 10000 optimization iterations. This is shown in Fig. \ref{fig:qualitative}. We observe that a few unlikely samples at peak 4, can lead to overconfident uncertainty estimates that prune the peak from the search space. Due to the reliance on the Wilson Score CI to provide reliable confidence estimates even for a low number of samples, this rarely happens in the WS-KDE, leading to a stable convergence toward the global maximum.

\subsection{Trap Optimization for Vibratory Feeders}
For the specific tasks of vibratory feeder design, the goal is to design traps that output a stream of parts in the same stable pose. Some tasks can constrain the trap output to deliver a part in a specific desired pose, $\mathcal{P}_{target}$, but when no such constraint exists the optimal trap configuration is one that outputs as narrow a stable pose distribution as possible. 
The proposed extension to WS-KDE enables this evaluation function to be applied to the trap design problem where an optimization task previously was limited by the requirement to provide a binomial outcome. In effect, this means that to start an optimization task, a designer would be forced to explicitly label a stable pose as the desired outcome of the trap and then for every iteration of the optimization, it would result in a success only if the part ended in this stable pose. When the dynamic behavior of parts paired with a trap is unpredictable, this domain knowledge can be hard to encode and an exhaustive list optimization (e.g. with each stable pose as the target) is the only option that ensures optimality in the design. Furthermore, the overall computational requirements scale proportionally with the number of discrete stable poses, which for complex part geometries can exceed 30. This problem can be overcome by defining the evaluation function $R(x)$ as the maximum frequency of occurrence of an outcome stable pose divided by the number of parts that were conveyed past the trap. So if ten parts convey past the trap and six of them end in the same stable pose this would result in $R(x)=0.6$. 

\begin{figure}
     \centering
     \includegraphics[width=0.95\linewidth]{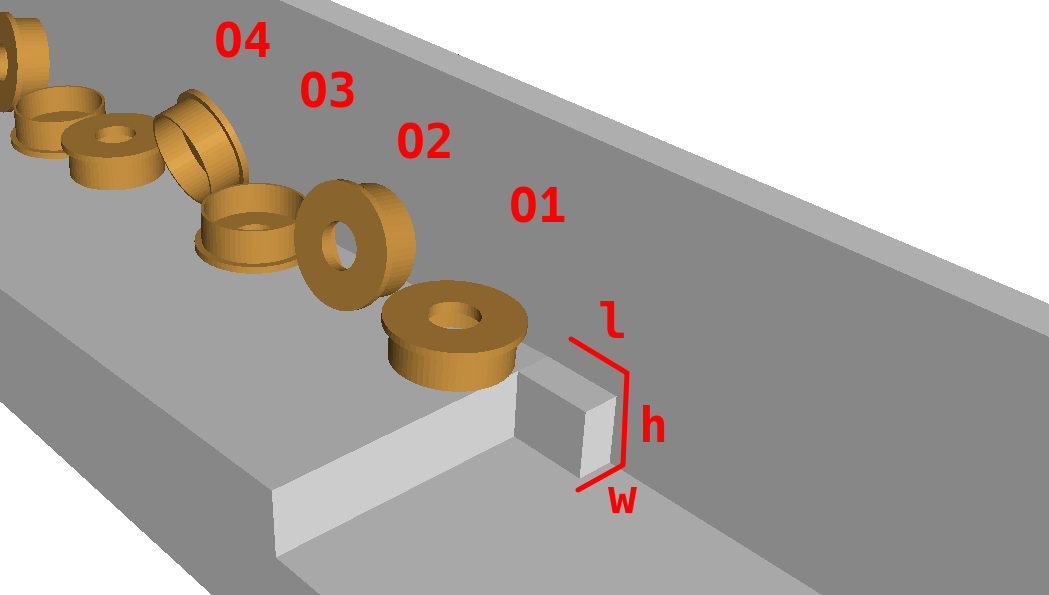}
     \caption{Experimental setup showing the parts on the trap-type being optimized. The three trap parameters are illustrated together with the four stable poses the cap part can assume.}
     \label{fig:expSetup}
 \end{figure}

For this trap design experiment, we demonstrate the proposed WS-KDE extension on a trap with three parameters and a cap-like part. These are shown in Fig.~\ref{fig:expSetup}. The trap is a reorientation type trap where the intended influence on the part is to topple it when it falls from the step. The step can assume the shape of a protruded surface from the highest point of elevation enabling parts to topple both parallel and perpendicular to the conveying direction. The three parameters of the trap are: the step height ($h$), the length ($l$), and the width ($w$) of the protruded surface. The part has an outer diameter of $19.8mm$, a height of $7.0mm$, a mass of $5.8g$, and an estimated center of gravity displaced $1.7mm$ towards the flat disc-like surface from its reference frame placed in the center of its geometry. The simulation parameters have been tuned based on test experiments with a real-world setup to minimize the reality gap.
%and enabled us to obtain realistic results as in~\cite{mathiesen2018optimisation}.\thmi{Måske reformulér så det ikke lyder somom vi har test i virkeligheden.}

% The distribution of stable poses prior to the trap is obtained from simulation:
% \begin{align}
% p \left( \{O1,O2,O3,O4 \} \right) = \{0.25,0.13,0.40,0.22\}
% \end{align}
The distribution of stable poses prior to the trap is obtained from simulation, $p \left( \{O1,O2,O3,O4 \} \right) = \{0.25,0.13,0.40,0.22\}$,
and the parameter ranges of the trap were chosen based on the part's dimensions (see Table~\ref{tab:trapParamsRanges}).
\begin{table}
    \renewcommand{\arraystretch}{1.3}
    \caption{The parameter ranges for the traps with min., max., and increment size of the discretized solution space. Units are in $[mm]$.}
    \label{tab:trapParamsRanges}
    \centering
    \begin{tabular}{l|ccc}
        Param. name  & $l$ & $h$  & $w$ \\
       \hline
        Min. & 2.0 & 2.0 & 2.0 \\
        Max. & 22.0 & 12.0 & 12.0 \\
        Inc. & 5.0 & 1.0 & 1.0 \\
    \end{tabular}

\end{table}

A simulation consists of the parts being spawned on the feeder track prior to the trap, conveying them past the trap and computing the reward $R(x)$ from the distribution of output stable poses. The optimization uses the same pruning and sampling strategy as the previous experiment. The simulation uses Open Dynamics Engine~\cite{ode} implemented through the simulation framework RobWorkSim~\cite{jorgensen2010robworksim} providing efficient and accurate contact generation required for simulating vibratory feeders.
We compare the results of the optimization performed using the proposed novel evaluation function with the results of an exhaustive optimization based on binomial outcomes. The simulations are run with multiple parts to capture realistic part interaction, although for the current implementation of the simulator the simulation time scales significantly with the number of parts. When simulations are evaluated the first and last part are disregarded as we assume a continuous stream of parts. The binomial outcome experiment simulations are run with three parts while the non-binomial outcome experiment is run with twelve parts which allows for a discretization step of the evaluation function of $0.1$. 

The kernel size for the experiment is chosen to ensure smoothing in the parameter space with an overlap for neighboring samples at $1\sigma$ resulting in $H=Diag\{5.0,1.0,1.0\}$. The optimization is run until the sample with the highest lower bound has a confidence interval $\hat \sigma_{ws,kde} \leq \pm0.05$. This can be considered a reliable result based on our experience with the accuracy of the simulation.
The results of the five optimizations are presented in Table~\ref{tab:trapresults}
\begin{table}
    \renewcommand{\arraystretch}{1.3}
    \caption{The results of the five optimizations identified by their target pose. From left to right: the estimated mean performance, the iteration number when reaching the confidence threshold, the pruning fraction at this iteration number, and parameter values of this sample $[mm]$.}
    \label{tab:trapresults}
    \centering
    \begin{tabular}{l|ccccccc}
         $\mathcal{P}_{target}$ & $ \hat{p}_{ws,kde}$ & Iter. & $I_{total}$   & $l_{best}$ & $h_{best}$ & $w_{best}$ &  \\
         \hline
         All & 0.73 & 4117  & 0.83 & 12 & 12 & 2 & \\
         O1  & 0.45 & 1989  & 0.69 & 17 & 3 & 3 & \\
         O2  & 0.14 & 2752 & 0.62 & 22 & 2 & 10 & \\
         O3  & 0.64 & 2725 & 0.54 & 7 & 11 & 6 & \\
         O4  & 0.19 & 2974  & 0.56 & 17 & 12 & 12 & \\
    \end{tabular}
\end{table}

From the results, it can be seen that our defined threshold for reliable estimates is reached at 4117 iterations for our optimization with no specific $P_{target}$. This must be seen in contrast to the exhaustive optimization totaling 10440 iterations to reach the target across the four tasks. It can also be seen that the pruning strategy has reduced the remaining sample space for samples which could potentially be better to $17\%$ compared to an average $40\%$ across the four binomial optimization tasks. This leaves us with significantly more information on the optimality of the trap design having spent $<40\%$ of optimization iterations. In addition, we emphasize that many parts have much more than $4$ stable poses.

Investigating the result with a dense sampling in the best parameter set shows that the solution found by the open-ended ($All$) optimization converges on a solution that always brings the majority of the parts into the $O3$ stable pose. However, the two optimizations do not find the same solution to be optimal. To explore this we ran the $O3$ optimization for 10000 iterations, observing that the parameter set found by the $All$ optimization was ranked at 67th best with a performance of $[\mu\pm CI] = [0.58\pm0.03]$. Considering there is no overlap in the estimated performance this suggests that the dynamics created by having twelve parts in the simulation is different than for only three. It is reasonable to assume that this benefits the accuracy of the estimated performance positively as a real-world feeder would contain hundreds of parts at a time and convey them in a steady stream. The proposed WS-KDE extension enables reaching certainty with fewer optimization iterations. This effectively allows us to prioritize using the available computational resources on simulation quality instead of quantity of simulations. This was evident in the non-binomial outcome experiment, which benefited from simulating 12 rather than 3 parts.

\section{Conclusion} 
\label{sec:Conclusion}
In this work, we have proved that the confidence bounds provided by the Wilson Score Kernel density estimator (WS-KDE) are applicable as bounds to any stochastic function with an output distribution with support on the closed interval [0;1] regardless of the distribution. Through simulation experiments, the coverage and convergence properties of WS-KDE have been evaluated in the context of Bayesian optimization and found to outperform standard KDE. Finally, we have shown that this newfound property of WS-KDE facilitates the use of a novel non-binary reward function for automating the design of vibratory part feeder traps, resulting in an optimal solution in fewer iterations.

It was mentioned in Sec. \ref{subsec:simulation} that coverage becomes less than the set confidence when the number of experiments becomes large. It has been derived~\cite{hardle2004nonparametric} that the theoretic bias in the density increases with increasing function curvature and increasing kernel size. Since the size of the confidence bounds decreases as the number of data points increases, but the bias remains unaffected, it is clear that coverage will decrease as the number of data points increases. The choice of kernel size is thus an essential parameter to choose wisely. We have performed initial studies of how good kernel sizes can be chosen based on qualitative assessments of $S(x)$ and data-driven approaches. This is ongoing work that will be reported in future papers.

%The theoretic density bias, derived in \cite{hardle2004nonparametric} as $Bias\{ \hat{f_h} (x)\} = \frac{h^2}{2}f^{''}\mu^2(K) + o(h^2)$ impacts robustness for a large number of experiments. The bias in the function estimate increases with increasing function curvature and increasing kernel size. Since the size of the confidence bounds decrease as the number of data points increases, but the the bias remains unaffected, it is clear that coverage will decrease as the number of data points increases. The choice of kernel size, is thus an essential parameter to choose wisely. We have performed initial studies of how good kernel sizes can be chosen based on qualitative assessments of $S(x)$ and also using data driven approaches. This is ongoing work that will be reported in future papers.
\section*{Appendix: Implementation}
The equations and simulations in this work have been presented in the context of 1D function estimation. However, the equations are, as noted in \cite{sorensen2020wilson}, straightforward to extend to multiple dimensions by following the derivations in \cite{sorensen2020wilson} using the standard  multivariate  kernel density equations provided in \cite{hardle2004nonparametric}. For ease of implementation, the multivariate versions of the WS-KDE equations are provided here.

% Note that all methods are implemented for batches, so e.g. when the equation takes an input $\x \in \mathbb{R}^d$, the actual implementation assumes $\x \in \mathbb{R}^{b\times d}$, where the given equation will be computed b times.

\begin{align}
    % &d: \text{dimensionality}\\
    % &n: \text{number of training samples}\\
    % &H = \Sigma ^ {1/2} \quad\text{where } \Sigma \text{ is the Kernel covariance}\\
    % &H = [\eigv_1, ...,\eigv_d] \text{diag}(\lambda_1,...,\lambda_d) [\eigv_1, ...,\eigv_d]^{-1}\\
    % &H^{-1/2} = [\eigv_1, ...,\eigv_d] \text{diag}(\lambda_1^{-1/2},...,\lambda_d^{-1/2}) [\eigv_1, ...,\eigv_d]^{-1}\\
    &k_{22} = \frac{1}{2^d\sqrt{\pi ^d}}\\
    &p_\text{wskde}(\x,z) = \frac{1}{n_h(\x) + z^2} \left(n_h(\x)m_h(\x)+\frac{z^2}{2}\right)\\
    &n_h(\x) = \frac{\Det{H}}{k_{22}}\sum_{i=0}^n K_{h,\x_i}(\x)\\
    &\sigma_\text{wskde}(\x,z) = \frac{z}{n_h(\x) + z^2}\nonumber\\
    &\quad\times\sqrt{n_h(\x)m_h(\x)(1-m_h(\x))+\frac{z^2}{4}}
\end{align}

where $d$ is the dimensionality of the search space, $n$ is the number of training samples, and $H = \Sigma ^ {1/2}$ is the bandwidth matrix with $\Sigma$ being the desired covariance of the Gaussian kernel.

The following equations are computed for every i'th training data point $\x_i$.\\
\begin{align}
    &K_i(\mathbf{u}_i) = (2\pi)^{-d/2}\exp ^{-\frac{1}{2}\mathbf{u}_i^\intercal\mathbf{u}_i}\\\
    &K_{h,\x_i}(\x) = \frac{1}{\Det{H}}K_{i}(H^{-1}(\x-\x_i))\\
    &m_h(\x) = \frac{\sum_{i=0}^n \mathbf{y}_i K_{h,\x_i}}{\sum_{i=0}^n K_{h,\x_i}}
\end{align}

Note that to ensure numerical stability, the equation $n_h(\x) = \frac{\Det{H}}{k_{22}}\sum_{i=0}^n K_{h,\x_i}(\x)$ is used instead of $n_h(\x) = \frac{n\Det{H}}{k_{22}}f_h(\x)$ and $f_h(\x) = \frac{1}{n}\sum_{i=0}^n K_{h,\x_i}(\x)$.


%%%%%%%%%%%%%%%%%%%%%%%%%%%%%%%%%%%%%%%%%%%%%%%%%%%%%%%%%%%%%%%%%%%%%%%%%%%%%%%%

%%%%%%%%%%%%%%%%%%%%%%%%%%%%%%%%%%%%%%%%%%%%%%%%%%%%%%%%%%%%%%%%%%%%%%%%%%%%%%%%



% \section*{ACKNOWLEDGMENT}
% We should discuss this. It is at least NOVO and MADE, but unclear how happy either of them are to have the other included.

% {\small
\bibliographystyle{IEEEtran}
\bibliography{references}
% }

\end{document}